\relax
\documentclass[11pt]{article}
\usepackage{acl}  
\usepackage{times}  
\usepackage{latexsym}
\usepackage[T1]{fontenc}
\usepackage[utf8]{inputenc}
\usepackage{microtype}
\usepackage{graphicx} 
\usepackage{natbib}  
\usepackage{caption} 
\DeclareCaptionStyle{ruled}{labelfont=normalfont,labelsep=colon,strut=off} 
\usepackage{algorithm}
\usepackage{algorithmic}
\usepackage{verbatimbox}
\usepackage{tikz}
\usetikzlibrary{positioning}
\definecolor{DarkGreen}{rgb}{0.0, 0.5, 0.0}
\usepackage{pgfplots}
\usepgfplotslibrary{groupplots,dateplot}
\usetikzlibrary{patterns,shapes.arrows}
\pgfplotsset{compat=newest}
\usepackage{silence}
\usepackage{newfloat}
\usepackage{listings}
\floatstyle{ruled}
\newfloat{listing}{tb}{lst}{}
\floatname{listing}{Listing}

\title{Maximum Bayes Smatch Ensemble Distillation for AMR Parsing}

\author{
Young-Suk Lee$\dagger$, Ramón Fernandez Astudillo$\dagger$, Thanh Lam Hoang$\ddagger$, 
  \\  
  \textbf{Tahira Naseem$\dagger$, Radu Florian$\dagger$,  Salim Roukos$\dagger$} \\
 \texttt{\{ysuklee,tanseem,raduf,roukos\}@us.ibm.com} \\
 \texttt{ramon.astudillo@ibm.com} \\ \texttt{t.l.hoang@ie.ibm.com}\\
 IBM Research AI$\dagger$, IBM Research - Ireland$\ddagger$
  }

\begin{document}
\maketitle
\begin{abstract}
AMR parsing has experienced an unprecendented increase in performance in the last three years, due to a mixture of effects including architecture improvements and transfer learning. Self-learning techniques have also played a role in pushing performance forward. However, for most recent high performant parsers, the effect of self-learning and silver data augmentation seems to be fading. In this paper we propose to overcome this diminishing returns of silver data by combining 
Smatch-based ensembling techniques with ensemble distillation.
In an extensive experimental setup, we push single model English parser performance to a new state-of-the-art, 85.9 (AMR2.0) and 84.3 (AMR3.0),  and return to substantial gains from silver data augmentation. 
We also attain a new state-of-the-art for cross-lingual AMR parsing for Chinese, German, Italian and Spanish. Finally we explore the impact of the proposed technique on domain adaptation, and show that it can produce gains rivaling those of human annotated data for QALD-9 and achieve a new state-of-the-art for BioAMR.
\end{abstract}

\section{Introduction}
\label{sec:intro}



Adoption of the Transformer architecture \cite{vaswani2017} for Abstract Meaning Representation (AMR) parsing \citep{cainlam2020,astudillo-etal-2020-transition} as well as pretrained language models \cite{bevilacqua2021aaai,zhou2021emnlp, bai-etal-2022-graph} have enabled an improvement of above $10$ Smatch points \cite{cai-knight-2013-smatch}, the standard metric, in the last two years. 

Data augmentation techniques have also shown great success in pushing the state-of-the-art of AMR parsing forward. These include generating silver AMR annotations with a trained parser \cite{konstas-etal-2017-neural, van2017neural}, with multitask pre-training and fine-tuning \cite{xu2020improving} as well as combining AMR to source text and silver AMR generation \cite{lee-etal-2020-transition} and stacked pre-training of silver data from different models -- from low performance to high performance silver data \citep{xia2021emnlpfindings}.
However, the latest BART-based state-of-the-art parsers, have shown diminishing returns for data augmentation. Both SPRING \cite{bevilacqua2021aaai} and Structured-BART \cite{zhou2021emnlp} gain a mere $0.5$ Smatch from self-learning, compared with over $1$ point gains of the previous, less performant, models. Since performance scores are already above where inter annotator agreement (IAA) is assumed to be, i.e. 83 for newswire and 79 for web text reported in \citep{banarescu-etal-2013-abstract}, one possible explanation is that we are reaching some unavoidable performance plateau.

In this work we show that we can achieve significant performance gains close to 2 Smatch point with the newly proposed data augmentation technique, contrary to the results from the previous state-of-the-art systems. The main contributions of this paper are as follows:


\begin{itemize}
\item We propose to combine  Smatch-based model ensembling \cite{barzdins2016,graphensemble2021} and ensemble distillation \citep{hinton2015distilling} of heterogeneous parsers to produce high quality silver data.

\item We offer a Bayesian ensemble interpretation of this technique as alternative to views such as Minimum Bayes Risk decoding \cite{goel2000minimum} and name the technique Maximum Bayes Smatch Ensemble (MBSE).

\item Applied to English monolingual parsing, MBSE distillation yields a new single system state-of-the-art (SoTA) on AMR2.0 (85.9) and AMR3.0 (84.3) test sets. 

\item Trained with Structured-mBART\footnote{\url{ https://github.com/IBM/transition-amr-parser/tree/structured-mbart}}, it yields new SoTA for Chinese (63.0), German (73.7), Italian (76.1) and Spanish (77.1) cross-lingual parsing. 

\item Applied to domain adaptation, MBSE distillation achieves the performance comparable to human annotations of QALD-9 training data and achieves new SoTA on BioAMR test set. 

\item We release QALD-9-AMR treebank\footnote{\url{https://github.com/IBM/AMR-annotations}}~at, which comprises 408 training and 150 test sentences.

\end{itemize}



%
%

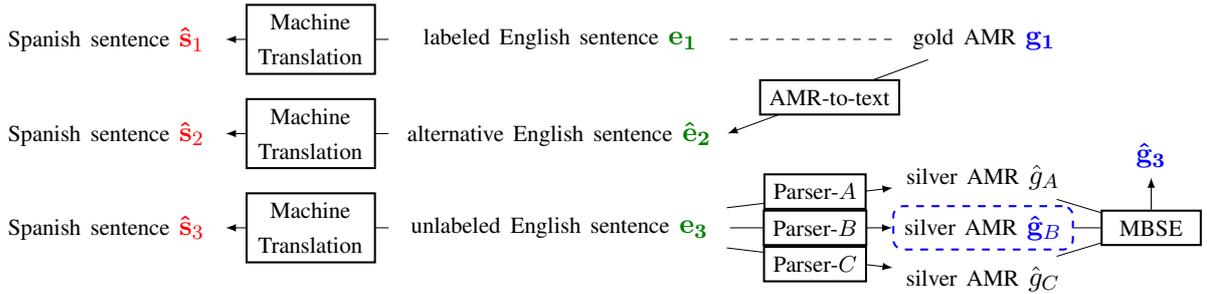
\begin{figure*}[!ht]
    \centering
\resizebox{1.01\textwidth}{!}{
\begin{tikzpicture}

\def \pxa {0}
\def \pxb {6.2}
\def \pxc {12}
\def \pxd {14.3}
\def \pya {0}
\def \pyb {-1.3}
\def \pyc {-2.6}
\def \pyca {\pyc + 0.7}
\def \pycb {\pyc + 0.0}
\def \pycc {\pyc - 0.7}

\node [text width=30mm,align=center] (ess) at (\pxa, \pya) {{\footnotesize Spanish sentence} {\color{red} $\mathbf{\hat{s}}_1$}};
\node [text width=44mm,align=center] (ens) at (\pxb, \pya) {{\footnotesize labeled English sentence} ${\color{DarkGreen} \mathbf{e_1}}$};
\node [text width=22mm,align=center] (ena) at (\pxc, \pya) {{\footnotesize gold AMR} {\color{blue} $\mathbf{g_1}$}};
\draw [latex-] (ess) to node [above] {} node [below] {} (ens);
\node [draw,thick,fill=white, text width=15mm,align=center] (parser1) at (\pxa + 2.8, \pya) {\footnotesize Machine Translation};
\path (ens) edge[-,draw,dashed] (ena);

\node [text width=30mm,align=center] (ess2) at (\pxa, \pyb) {{\footnotesize Spanish sentence} {\color{red} $\mathbf{\hat{s}}_2$}};
\node [text width=44mm,align=center] (ens2) at (\pxb, \pyb) {{\footnotesize alternative English sentence} ${\color{DarkGreen} \mathbf{\hat{e}_2}}$};
\draw [latex-] (ess2) to node [above] {} node [below] {} (ens2);
\node [draw,thick,fill=white, text width=15mm,align=center] (parser1) at (\pxa + 2.8, \pyb) {\footnotesize Machine Translation};
\draw [latex-] ({\pxb + 2.3}, \pyb) to node [above] {} (ena);
\node [draw,thick,fill=white] (amrtotxt) at (\pxb + 3.7, \pyb + 0.5) {\footnotesize AMR-to-text};

\node [text width=30mm,align=center] (ess3) at (\pxa, \pyc) {{\footnotesize Spanish sentence} {\color{red} $\mathbf{\hat{s}}_3$}};
\node [text width=44mm,align=center] (ens3) at (\pxb, \pyc) {{\footnotesize unlabeled English sentence} ${\color{DarkGreen} \mathbf{e_3}}$};
\node [text width=22mm,align=center] (ensa0) at (\pxc, \pyca) {{\footnotesize silver AMR} $\hat{g}_A$};
\node [draw=blue, text width=22mm,align=center, rounded corners, dashed, thick] (ensa1) at (\pxc, \pycb)  {{\footnotesize silver AMR} {\color{blue} $\mathbf{\hat{g}}_B$}};
\node [text width=22mm,align=center] (ensa2) at (\pxc, \pycc)  {{\footnotesize silver AMR} $\hat{g}_C$};
\draw [latex-] (ess3) to node [above] {} node [below] {} (ens3);
\node [draw,thick,fill=white, text width=15mm,align=center] (parser1) at (\pxa + 2.8, \pyc) {\footnotesize Machine Translation};
\draw [-latex] (ens3) to node [above] {} (ensa0);
\draw [-latex] (ens3) to node {} (ensa1);
\draw [-latex] (ens3) to node [below] {} (ensa2);
\node [draw,thick,fill=white] (parser1) at (\pxb + 3.5, \pyca - 0.2) {\footnotesize Parser-$A$};
\node [draw,thick,fill=white] (parser1) at (\pxb + 3.5, \pycb) {\footnotesize Parser-$B$};
\node [draw,thick,fill=white] (parser1) at (\pxb + 3.5, \pycc + 0.2) {\footnotesize Parser-$C$};
\node [draw,thick,text width=11mm,align=center] (reranking) at (\pxd, \pyc) {\footnotesize MBSE};
\draw (ensa0) -- (reranking);
\draw (ensa1) -- (reranking);
\draw (ensa2) -- (reranking);

\node [align=center] (mbseout) at (\pxd, {\pyc + 1}) {\color{blue} $\mathbf{\hat{g}_3}$};
\draw [-latex] (reranking) to node {} (mbseout);

\end{tikzpicture}
}

\caption{Data augmentation framework: Given a labeled example in English $({\bf \color{DarkGreen} e_1}, {\bf \color{blue} g_1})$, we use an AMR-to-text generation system to generate an alternative input text ${\bf \color{DarkGreen} \hat{e}_2}$ for ${\bf \color{blue} g_1}$ following \cite{lee-etal-2020-transition}. Given a sentence ${\bf \color{DarkGreen} e_3}$, and various state-of-the-art off-the-shelf parser outputs ($A$, $B$, $C$), Maximum Bayes Smatch Ensemble (MBSE) produces a single annotation for each input sentence by selecting from existing AMRs or their modified versions. MBSE is only applied to unlabeled English sentences to produce ${\bf \color{blue} \hat{g}_3}$. Following \cite{damonte-cohen-2018-cross}, we translate the English sentences to e.g. Spanish, to yield new training samples $({\bf \color{red} \hat{s}_1}, {\bf \color{blue}g_1}), ({\bf \color{red} \hat{s}_2}, {\bf \color{blue}g_1}), ({\bf \color{red} \hat{s}_3}, {\bf \color{blue}\hat{g}_3})$ to train a Spanish cross-lingual parser. We use the English pairs $({\bf \color{DarkGreen} e_1}, {\bf \color{blue}g_1}), ({\bf \color{DarkGreen} \hat{e}_2}, {\bf \color{blue}g_1}), ({\bf \color{DarkGreen} e_3}, {\bf \color{blue}\hat{g}_3})$ to train an English parser. }
    \label{fig:AAAI2022_dataaugment}


\end{figure*}
\section{Maximum Bayes Smatch Ensemble}
\label{mbsd}


Ensemble distillation \cite{hinton2015distilling} integrates knowledge of different teacher models into a student model. For sequence to sequence models, e.g. machine translation, it is possible to ensemble models by combining probabilities of words given context at each time step \cite{kim2016emnlp,freitag2017arxiv}. Syntactic and semantic parsers model a distribution over graphs that is harder to integrate across teacher models in an optimal way. For particular cases like dependency parsing, it is possible to ensemble teachers based on the notion of edge attachment \cite{kuncoro2016emnlp}, which is related to the usual evaluation metric, Label Attachment Score (LAS). However, AMR graphs are quite complex and not explicitly aligned to words. The standard Smatch \cite{cai-knight-2013-smatch} metric approximates the NP-Complete problem of aligning nodes across graphs with a hill climbing algorithm. This illustrates the difficulty of achieving consensus across teachers for AMR ensembling. 

Prior work ensembling AMR graphs has leveraged Smatch directly or its hill climbing strategy for ensembling. The ensemble in \cite{barzdins2016} selects, among a number of candidate AMRs, the one that has the largest average Smatch with respect to all sampled AMRs. The ensemble in \cite{graphensemble2021}, modifies the candidate AMRs to increase consensus as measured by coverage. Then it selects from the union of original and modified graphs for the one with highest coverage or largest average Smatch.  
One possible intepretation of both techniques is that of Minimum Bayes Risk (MBR) decoding, a well established method in Automatic Speech Recognition (ASR) \cite{goel2000minimum} and Machine Translation (MT) \cite{kumar2004minimum}. Assuming that we have a model predicting a graph from an input sentence $p(g \mid w)$, normal decoding entails searching among model outputs $g$ for the one that has the highest likelihood according to the model $p(g \mid w)$. MBR searches instead for the model output that minimizes the risk with respect to the distribution of possible human (gold) outputs for a given input
\begin{eqnarray}
\hat{g} = \arg\min_g \{E_{p(g^h \mid w)} \{R(g, g^h)\}\}\nonumber
\end{eqnarray}
where $p(g^h \mid w)$ is the distribution of correct human outputs, e.g. given by multiple annotators, and $R$ is a risk function that measures how severe deviations from $g^h$ are. In this case risk would be minus \textrm{Smatch}. Since in practice $p(g^h \mid w)$ is not available, MBR takes often the strong assumption of replacing $p(g^h \mid w)$ by the model distribution itself $p(g \mid w)$. 

Here we suggest another Bayesian interpretation, that requires less strong assumptions than MBR, a Bayesian model ensemble \cite{wilson2020bayesian}. Indeed techniques above can be seen as solving
\begin{eqnarray}
\hat{g} = \arg\max_{g \in \mathcal{G}} \{E_{p(\mathcal{M} \mid \mathcal{D})} \{\mathrm{{\footnotesize Smatch}}(g, \tilde{g}_{\mathcal{M}})\}\}\nonumber
\end{eqnarray}
where $p(\mathcal{M} \mid \mathcal{D})$ is the distribution of models $\mathcal{M}$ given training data $\mathcal{D}$, approximated by a sample average of models of different architectures or different random seeds, and
\begin{eqnarray}
\tilde{g}_{\mathcal{M}} = \mathrm{post}\left(\arg\max_{y} \left\{\prod_{t=1}^{|y|}p_{\mathcal{M}}(y_t \mid y_{<t}, w)\right\}\right)\nonumber
\end{eqnarray}
is the output of a conventional decoding process for each parser prediction distribution $p_{\mathcal{M}}$, including post-processing $\mathrm{post}()$. This process differs across models indexed by $\mathcal{M}$, for example $y$ can be transition actions or linearized graphs and $\mathrm{post}()$ running the state-machine or linearized graph post-processing\footnotemark\footnotetext{We consider only auto-regressive models in this work but this approach could also encompass e.g. graph-based parsers.}. $\mathcal{G}$ is the space of candidate graphs, which in \citet{barzdins2016} are the AMRs resulting from decoding each sample from $p(\mathcal{M} \mid \mathcal{D})$ and in \citet{graphensemble2021} are those same graphs plus the modified pivot graphs. There is in principle no restriction on how to build the set $\mathcal{G}$. \textit{Decoding} a graph $g \in \mathcal{G}$ means here selecting the member of that set maximizing the expected Smatch and is different from each parser's decoding process. 

If we replace $\mathrm{Smatch}()$ by an indicator function on the decoding outputs $1_{g = \tilde{g}_{\mathcal{M}}}$, then 
\begin{eqnarray}
\hat{g} = \arg\max_{g \in \mathcal{G}} \{E_{p(\mathcal{M} \mid \mathcal{D})} \{1_{g = \tilde{g}_{\mathcal{M}}}\}\}\nonumber
\end{eqnarray}
recovers majority voting of AMR graphs. Since the space of graphs is exponentially large on the input size, this would be too sparse to attain meaningful vote counts. The propagation of the uncertainty in $p(\mathcal{M} \mid \mathcal{D})$ through the $\mathrm{Smatch}()$ transformation both solves the sparsity problem, and allows optimization on a space that is better related to the target metric. The method will be henceforth described here as Maximum Bayes Smatch Ensemble distillation (MBSE distillation). 

In what follows, we will consider three versions for ensembling, the \textrm{Smatch} version of \citet{graphensemble2021} (graphene-Smatch), the average-Smatch selection of \citet{barzdins2016}, and a greedy version of \citet{barzdins2016} where we select the two highest \textrm{Smatch} AMRs and from that pair, keep the graph with the highest \textrm{Smatch} with respect to the remaining graphs (greedy-select). The greedy-select algorithm is given in Algorithm~\ref{alg:algorithm1} of Appendix~\ref{sec:appendix_algorithm} and performs similarly to the average-Smatch of \citet{barzdins2016}.

\section{Silver Training Strategy}
\label{sec:silverstrategy}

We now describe the AMR silver training strategy proposed in this work. This strategy creates high quality English and cross-lingual AMR annotations for unlabeled data with MBSE and alternative input sentences of gold AMRs via AMR-to-text.

As depicted in Fig.~\ref{fig:AAAI2022_dataaugment}, we start with 1) a set of gold-labeled (English sentence, AMR) pairs, 2) a set of unlabeled English sentences and 3) pre-trained English-to-foreign language Machine Translation systems. Assuming $N$ off-the-shelf AMR parsers, we train each of the $N$ parsers using the gold data with their respective training procedure. 
More than one random seed may be trained for some parsers, leading to more than $N$ AMR parses for each input sentence. 

After the parsers have been trained, we use them to parse the unlabeled English text as in \citet{konstas-etal-2017-neural}. Interpreting the set of trained models as samples of the model distribution, we apply the MBSE distillation methods described in Section.~\ref{mbsd}. We apply all variations of the MBSE algorithms including graphene-Smatch, greedy-select and average-Smatch algorithms. 

For English parsers, the MBSE distilled AMR annotations are added to the human-annotated gold treebanks for enhanced model training. For cross-lingual parsers, we translate all English input sentences to the target foreign languages and train respective cross-lingual parsers with pairs of (Foreign language input sentences, AMR graphs in English), following \cite{damonte-cohen-2018-cross}.  

Following \citet{lee-etal-2020-transition},  we also apply an AMR-to-text model \cite{mager-etal-2020-gpt,ribeiro2020investigating,bevilacqua2021aaai}  to generate additional sentences for human-annotated AMR. We filter out the generated texts if they are too similar (BLEU $>$ 0.9) or too dissimilar (BLEU $<$ 0.1) to the original input texts, as measured by BLEU \citep{papineni2002acl}. AMR-to-text generation\footnote{We use \url{https://github.com/SapienzaNLP/spring}.} is used for cross-lingual AMR parser training only.

\begin{table*}[!t]
\centering
\begin{tabular}{l|l|rr||l|l|rr}
\hline
\multicolumn{4}{c}{\textbf{For Standard Experiments}} & \multicolumn{4}{c}{\textbf{For Domain Adaptation}} \\
\hline
\textbf{Dataset}  & \textbf{Split} & \textbf{Sents} & \textbf{Tokens}  & \textbf{Dataset}  & \textbf{Split} & \textbf{Sents} & \textbf{Tokens}\\
\hline 
AMR2.0   & Train  & 36,521 & 653K & QALD-9-AMR (\textit{new}) & Train  & 408 & 3,475\\
                    & Test   & 1,371 & 30K                 &                                   & Test   & 150 & 1,441 \\
                    & Dev.    &  1,368 & 29K               &                                   &&&\\ 
\hline
AMR3.0   & Train  & 55,635 & 1M   & Bio AMR  & Train  & 5,452 & 231K \\
                                  & Test   & 1,898 & 39K   &                                             & Test   & 500 & 22K  \\
                                  & Dev.   & 1,722 & 37K  &     &      &       & \\
\hline
                                  &        &       &      & LP  & Test & 1,562 & 21K\\
\hline
PropBank   & silver$^{1}$ & 20K & 386K     & SQuAD2.0-Q  & silver$^{q}$ & 135K & 1.5M  \\
SQuAD2.0-C    & silver$^{1}$ & 70K & 2M  & BioNLP-ST-2011  & silver$^{b}$ & 15K & 460K \\
Ontonotes5.0  & silver$^{2}$ & 59K & 1.1M  & CRAFT  & silver$^{b}$ & 27K & 740K\\
WikiText-103  & silver$^{3}$ & 70K & 2M      &&&&\\
\hline

\end{tabular}
\caption{Corpus statistics for the standard benchmark experiments on AMR2.0 and AMR3.0 test sets (left) and domain adaptation experiments (right).  Silver indicates the unlabeled data for silver training.}
\label{tab:benchmarkdata}
\end{table*}

\section{Experimental Setup}
\label{experimentalsetup}

\subsection{Corpus Statistics and QALD-9-AMR} 
Table~\ref{tab:benchmarkdata} details the corpora considered for the standard benchmark experiments on AMR2.0 and AMR3.0 test sets (lef) and out-of-domain data used for domain adaptation experiments (right). Silver indicates the unlabeled data for silver AMR acquisition.
SQuAD2.0-Q(uestions) are for QALD-9 (silver$^{q}$) and PubMed, BioNLP-2011 \citep{jindongkim2011} and CRAFT  \citep{cohen2017springer} for BioAMR (silver$^{b}$).


Since there were no human annotations of QALD-9 corpus, we created QALD-9-AMR treebank. QALD-9 training/test data have been annotated by 3 skilled resident human annotators with experience in AMR annotations over a year. Each of the annotators annotated both the train and test data sets, followed by cross validation by each other. The final annotations were adjudicated by the most experienced annotator. 
Inter-annotator agreement (IAA) rate on a subset of 158 training sentences is over 95\% in Smatch. The data is made publicly available under an Apache2 license. 

\begin{table*}
\centering
\begin{tabular}{l|c|c|c|c|c} \hline
Models            & AMR2.0 & AMR3.0 & Q9AMR & LP & BioAMR \\ \hline
APT~\citep{zhou2021naacl} & 83.0 & 81.1 & 83.7 & 79.0 & 55.2 \\
Structured-BART~\citep{zhou2021emnlp} & 84.6 & 83.1 & 87.7 & 81.0 & 62.4 \\
SPRING$_{1}$~\citep{bevilacqua2021aaai}      & 84.2 & 83.2 & 87.7 & 81.3 &  61.6 \\
SPRING$_{2}$ ~\citep{bevilacqua2021aaai}     & 83.8 & 82.9 & 86.4 & 81.0 &  60.5 \\
AMRBART~\citep{bai-etal-2022-graph} & 85.4 & 84.2 & 88.0 & 82.3 & 63.4 \\\hline
aver.-Smatch (A) ~\cite{barzdins2016} & 86.2 & 84.9 & 89.0 & 82.9 & 64.1 \\
graphene-Smatch (P) ~\cite{graphensemble2021}& \textbf{86.7} & \textbf{85.4} & \textbf{89.3} & \textbf{83.1} & \textbf{65.8} \\
greedy-select (G)  & 85.9 & 84.8 & 88.8 & 82.8 & 63.9 \\
\hline
\end{tabular}
\caption{English parsing performance in Smatch in general domain and domain adaptation for recent
state-of-the-art systems (top). Performance in Smatch for the ensemble of all systems using different Smatch-based ensembling techniques (bottom). SPRING$_{1}$ and SPRING$_{2}$ are $2$ random seeds of the same model. Highest scores are \textbf{boldfaced}.}
\label{tab:algorithmresults}
\end{table*}

\subsection{Parsing Models}

We use 4 off-the-shelf AMR parsers to parse un-annotated raw texts. We train the parsers following their standard configurations.

\textbf{APT}~\citep{zhou2021naacl}\footnote{\url{github.com/IBM/transition-amr-parser/tree/action-pointer}, Apache2 License} is a transition-based parser that combines hard attention over sentences with a target side action pointer mechanism to decouple source tokens from node representations and address alignments. 
Cross-attention of all decoder layers is used for action-source alignment. 

\textbf{SPRING}~\citet{bevilacqua2021aaai}\footnote{\url{github.com/SapienzaNLP/spring}, CC BY-NC-SA 4.0} fine-tunes BART \citep{lewis2019arxiv} to predict linearized AMR graphs, avoiding complex pipelines.

\textbf{Structured-BART}~\citet{zhou2021emnlp}\footnote{\url{github.com/IBM/transition-amr-parser}, Apache2 License} models the transition-based parser state within a pre-trained BART architecture, outperforming SPRING. This is the main parser for our work. 

\textbf{AMRBART}~\citet{bai-etal-2022-graph}\footnote{\url{https://github.com/muyeby/AMRBART}, MIT License} improves the structure awareness of pre-trained BART over AMR graphs by introducing node/edge denoising and sub-graph denoising tasks, for graph-to-graph pre-training, achieving significant improvement over previous BART-based systems.

\subsection{Structured-mBART} 
\label{structuredmbart}

For cross-lingual AMR parsing, we adapt Structured-BART by replacing the pretrained BART with mBART of \citep{mbart2020}, henceforth Structured-mBART. The codebase is made publicly available under an Apache2 license. Structured-mBART diverges from Structured-BART mainly in input processing and vocabulary:

\begin{itemize}
    \item For task vocabulary, Structured-mBART includes \textasciitilde{250K} sentencepiece tokens of \citep{kudo2018acl} including 25 language tags, e.g.  es\textunderscore XX, whereas Structured-BART includes \textasciitilde{50K} BPE tokens of \citep{sennrich2016acl}.
    \item We append the source language tag to the end of each input sentence without specifying the target language tag for Structured-mBART.
    \item For Structured-mBART, we set the learning rate to $3\mathrm{e}{-5}$, cf. $1\mathrm{e}{-4}$ of Structured-BART, and  move the layer normalization to the beginning of each transformer block. 
\end{itemize}

We obtain contextualized embeddings from the pre-trained mBART for multilingual input sentence representations. For  target action sequences, we map the sentencepiece
tokens to the corresponding  target token, by averaging all values from the sentencepiece tokens corresponding to the target token. For German, Italian and Spanish input texts, we apply the tokenizer from JAMR  parser\footnote{\url{https://github.com/jflanigan/jamr}} before sentencepiece tokenization. For Chinese, we directly apply the sentencepiece tokenizer.

\section{Results}
\label{sec:results}

To explore the effect of the proposed MBSE distillation and training strategy, we consider an extensive experimental setup
including standard English benchmarks (Section~\ref{sec:englishbenchmark}), cross-lingual benchmarks (Section \ref{crosslingual}) and out of domain data sets (Section \ref{domainadaptation}).\footnote{We also applied the technique to APT, observing similar performance gains when using MBSE distillation.} 
For model training and selection details, see Appendix~\ref{sec:appendix_param} and Appendix~\ref{sec:appendix_implementation}. 

We first provide the performance evaluation of each ensembling technique used in MBSE in Table~\ref{tab:algorithmresults} to demonstrate the effectiveness of the ensemble techniques by themselves. We test the algorithm on the standard test data sets from AMR2.0 and AMR3.0 and three out-of-domain data sets, Q9AMR (QALD-9-AMR), LP (Little Prince) and BioAMR in  Table~\ref{tab:benchmarkdata}.
We consider here only standard English AMR parsing. As expected, all MBSE algorithms, average-Smatch, graphene-Smatch and greedy-select,  improve individual models by large margins. Note that while the ensembles outperform single model state-of-the-art by a large margin, the use of heterogeneous ensembles of models is computationally prohibitive in practice, both due to the cost of running different models but also the ensembling techniques.

\subsection{English AMR Parsing}
\label{sec:englishbenchmark}

\begin{table*}[t]
\centering
\begin{tabular}{l|c|cc|cc}
\hline 
 \textbf{Models}                                    & silver &\multicolumn{2}{c|}{\textbf{AMR2.0}} & \multicolumn{2}{c}{\textbf{AMR3.0}} \\ \hline
 \citet{tnaseem2019}                                &        & \multicolumn{2}{c|}{75.5} & \multicolumn{2}{c}{-}\\
 \citet{zhang2019arxiv}                             &        & \multicolumn{2}{c|}{76.3{\small{$\pm$0.1}}} &\multicolumn{2}{c}{-} \\
 \citet{zhang2019emnlp}                             &        & \multicolumn{2}{c|}{77.0{\small{$\pm$0.1}}} & \multicolumn{2}{c}{-}\\
 \citet{cainlam2020}                                &        & \multicolumn{2}{c|}{80.2} & \multicolumn{2}{c}{-}\\
 \citet{astudillo-etal-2020-transition}             &        & \multicolumn{2}{c|}{80.2{\small{$\pm$0.0}}} & \multicolumn{2}{c}{-}\\
 \citet{lyu2020}                                    &        & \multicolumn{2}{c|}{-}                      & \multicolumn{2}{c}{75.8} \\
 \citet{lee-etal-2020-transition}                   &   85K  & \multicolumn{2}{c|}{81.3{\small{$\pm$0.0}}} & \multicolumn{2}{c}{-}\\
 \citet{xu2020improving}                            &  14M   & \multicolumn{2}{c|}{81.4} & \multicolumn{2}{c}{-}\\
 \citet{bevilacqua2021aaai}                         & 200K   & \multicolumn{2}{c|}{84.5} & \multicolumn{2}{c}{83.0} \\
 \citet{zhou2021naacl}                              &  70K   & \multicolumn{2}{c|}{82.6{\small{$\pm$0.1}}} & \multicolumn{2}{c}{80.3}  \\
 \citet{xia2021emnlpfindings}                       &  1.8M  &  \multicolumn{2}{c|}{84.2} & \multicolumn{2}{c}{-} \\
\citet{bai-etal-2022-graph}                         &  200K  & \multicolumn{2}{c|}{\textbf{85.4}} & \multicolumn{2}{c}{\textbf{84.2}} \\\hline

\citet{zhou2021emnlp}                               &        & sep-voc & joint-voc & sep-voc & joint-voc \\
Structured-BART-baseline                            &        & 84.0\small{$\pm$0.1} & 84.2\small{$\pm$0.1} & 82.3\small{$\pm$0.0}  & 82.0\small{$\pm$0.0} \\ 
+ self-trained silver$^{1}$                         &  90K   &  -   & 84.7\small{$\pm$0.1} & 82.7\small{$\pm$0.1} & 82.6\small{$\pm$0.0} \\
+ self-trained silver$^{1}$ + ensemble dec.      &  90K   & - & 84.9 &  83.1 & - \\
\hline
\textbf{Ours below} (Struct-BART)                   &        & sep-voc & joint-voc & sep-voc & joint-voc \\
+ SPRING silver$^{1}$                               &  90K   & 84.8\small{$\pm$0.1} & 84.8\small{$\pm$0.0} & 83.0\small{$\pm$0.0} & 83.2\small{$\pm$0.1} \\
+ SPRING + self-trained silver$^{1}$ (50:50)        &  90K   & 84.8\small{$\pm$0.1} & 84.7\small{$\pm$0.0} & 83.0\small{$\pm$0.0} & 83.2\small{$\pm$0.1} \\\hline

\multicolumn{6}{l}{Ensemble-4 distillation (APT + Structured-BART + SPRING$_{1}$ + SPRING$_{2}$)} \\\hline
+ MBSE-P silver$^{1}$                 &  90K   & 85.1\small{$\pm{0.1}$} & 85.1\small{$\pm{0.1}$} &  83.2\small$\pm{0.1}$ &  83.5\small$\pm{0.1}$ \\
+ MBSE-G silver$^{1}$                               &  90K   & 85.0\small{$\pm$0.0} & 85.2\small{$\pm$0.1} & 83.4\small{$\pm$0.0} & 83.5\small{$\pm$0.0} \\
+ MBSE-G silver$^{1+2}$                             &  149K  & 85.3\small{$\pm{0.1}$} & 85.4\small{$\pm{0.1}$} & 83.6\small{$\pm{0.1}$} & 83.7\small{$\pm{0.1}$} \\
+ MBSE-G siver$^{1+2+3}$ &  219K  & 85.3\small$\pm{0.1}$     & 85.5\small$\pm{0.1}$ & 83.7\small$\pm{0.0}$ & 83.9\small$\pm{0.0}$\\
+ MBSE-G silver$^{1+2+3}$ + ensemble dec.           &  219K  & \textbf{85.6} & \textbf{85.7} & \textbf{84.0} & \textbf{84.2} \\
\hline
\multicolumn{5}{l}{Ensemble-5 distillation (APT + Structured-BART + SPRING$_{1}$ + SPRING$_{2}$ + AMRBART)} \\\hline
+ MBSE-A silver$^{1}$                               &  90K   &  & 85.3\small{$\pm$0.1} &  & 83.6\small{$\pm$0.1} \\
+ MBSE-A silver$^{1+2}$                             & 149K   &  & 85.5\small{$\pm$0.0} & & 84.0\small{$\pm$0.0} \\
+ MBSE-A silver$^{1+2+3}$                           & 219K   &  & 85.7\small{$\pm$0.0} &  & 84.1\small{$\pm$0.0} \\
+ MBSE-A silver$^{1+2+3}$ + ensemble dec.           & 219K   & & \textbf{85.9}  &  & \textbf{84.3} \\\hline
\end{tabular}
\caption{Smatch scores on AMR2.0 and AMR3.0 test data. 
Lower rows show Structured-BART performances with various silver data augmentations. sep-voc denotes separate vocabulary and joint-voc, joint vocabulary. The numbers prefixed by $\pm$ indicate the standard deviation of Smatch scores across 3 seeds.
}
\label{tab:structbart} 
\end{table*}

As displayed in Table~\ref{tab:benchmarkdata}, we consider the standard AMR2.0 (\textsc{LDC2017T10}) and AMR3.0 (\textsc{LDC2020T02}) treebank as gold data. For ensemble distillation, we use the data sets denoted by silver$^{1}$ for comparison with previous work, and silver$^{2}$ and silver$^{3}$ to investigate the impact of unlabeled corpus size on model performance. For silver$^{1}$, we use all sentence examples in PropBank (\textsc{LDC2004T14}). From SQuAD2.0-C(ontexts)\footnote{ \url{https://rajpurkar.github.io/SQuAD-explorer/}} we filter out the \textasciitilde{92K} sentences, removing bad utf8 encoding (\textasciitilde{7K}) and ill-formed disconnected graphs produced by APT (\textasciitilde{15K}). Silver$^{2}$ comprises Ontonotes5.0 (\textsc{LDC2013T19}) and silver$^3$ WikiText-103.

The results are shown in Table~\ref{tab:structbart}. The lower part of the table (denoted by \textbf{Ours}) compares the performances of Structured-BART in various silver data augmentation setups including our proposed MBSE distillation. With the same unlabeled corpus silver$^{1}$, greedy-select distillation improves 1.0 Smatch point on AMR2.0 (84.2 vs. 85.2) and 1.5 Smatch point on AMR3.0 (82.0 vs. 83.5) over the Structured-BART baselines. Graphene-Smatch distillation performs similarly to  greedy-select one. 

To isolate the effect of ensembling, we provide two additional baselines: 1) silver obtained from SPRING, which is expected to have complementary information to self-trained silver, and 2) an equal mixture of SPRING and Structured-BART (random 50:50), which tests if the MBSE selection strategy bears any effect. MBSE distillation outperforms these two baselines by between $0.2$ and $0.5$ Smatch point, depending on the scenario, proving that MBSE selection has a clear positive effect.

We also investigate the impact of unlabeled corpus size on model performance by adding silver$^{2}$ and silver$^{3}$ to silver$^{1}$, i.e. silver$^{1+2}$ and silver$^{1+2+3}$. We observe additional 0.3-0.4 improvement, complementary to the one obtainable with conventional ensemble decoding. This pushes the numbers to $85.7$ and $84.2$, setting a new SoTA for single system with 4 model ensemble (Ensemble-4) distillation. Note that using 5 model ensemble (Ensemble-5) distillation moves the Smatch scores even higher to 85.9 for AMR2.0 and 84.3 for AMR3.0. 

\begin{table*}[ht]
\centering
\resizebox{1.01\textwidth}{!}{
\begin{tabular}{l|c|cccc}
\hline
\textbf{Models} & \textbf{LM}  & \textbf{DE} & \textbf{ES} & \textbf{IT} & \textbf{ZH} \\
\hline
\multicolumn{6}{l}{\textbf{Translate and Parse Pipelines}}  \\  
\hline
{\citet{uhrig2021iwpt}} & & 67.6 & 72.3 & 70.7 & 59.1  \\
WLT+Structured-BART+MBSE-G silver$^{1}$ & BART & 73.9 & 76.5 & 76.1 & 63.7 \\
WLT+Structured-BART+MBSE-A silver$^{1+2+3}$ & BART & \textbf{74.6} & \textbf{77.1} & \textbf{76.7} & \textbf{64.0} \\
 \hline
\multicolumn{6}{l}{\textbf{Cross-lingual Parsers}} \\
\hline
{\citet{blloshmi-etal-2020-xl}} &   & 53.0 & 58.0 & 58.1 & 43.1 \\
{\citet{youngsuk2021eacl}} (85K silver AMR) & XLMR   & 62.7 & 67.9 & 67.4 & -- \\
{\citet{sgl2021naacl}} (5M parallel corpus) & mBART$_{mt}$  & 69.8 & 72.4 & 72.3 & 58.0   \\
{\citet{cai2021aclfinding}} &   & 64.0 & 67.3 & 65.4 & 53.7  \\
{\citet{xlpt2021acl}} &   & 70.5 & 71.8 & 70.8 & -- \\
{\citet{cai2021emnlpfindings}} (320K silver AMR) & mBARTmmt & \textbf{73.1} & \textbf{75.9} & \textbf{75.4} & \textbf{61.9} \\\hline
\textbf{Ours below} (with Structured-mBART) & & & & &    \\
Structured-mBART-baseline  & mBART  &   69.9\small{$\pm$0.0} & 74.4\small{$\pm$0.3} & 73.3\small{$\pm$0.2} & 59.9\small{$\pm$0.0} \\\hline
\multicolumn{6}{l}{Ensemble-4 distillation (APT + Structured-BART + SPRING$_{1}$ + SPRING$_{2}$)} \\\hline
+ MBSE-G silver$^{1}$   & mBART &   72.5\small{$\pm$0.1} & 76.5\small{$\pm$0.2} & 75.4\small{$\pm$0.0}  & 62.2\small{$\pm$0.1} \\
+ MBSE-G silver$^{1}$+AMR2Text & mBART &   72.9\small{$\pm$0.1} & 76.6\small{$\pm$0.0} & 75.6\small{$\pm$0.0} & 62.3\small{$\pm$0.0} \\
+ MBSE-G silver$^{1}$+AMR2Text + ens. dec. & mBART &  73.2 & 76.9 & 75.7 & 62.7 \\\hline
\multicolumn{6}{l}{Ensemble-5 distillation (APT + Structured-BART + SPRING$_{1}$ + SPRING$_{2}$ + AMRBART)} \\\hline
+ MBSE-A silver$^{1+2+3}$ & mBART & 73.5\small{$\pm$0.1} & \textbf{77.1}\small{$\pm$0.2} & 76.0\small{$\pm$0.1} & 62.7\small{$\pm$0.1} \\
+ MBSE-A silver$^{1+2+3}$ + ens. dec. & mBART & \textbf{73.7} & 77.0 & \textbf{76.1} & \textbf{63.0} \\\hline
\end{tabular}
}
\caption{\label{tab:Results_Test2.0}
Cross-lingual parser Smatch scores on AMR2.0 human translated test sets. mBART$_{mt}$ of \citet{sgl2021naacl} indicates the mBART model fine-tuned on both semantic parsing tasks and the MT data. mBARTmmt of \citet{cai2021emnlpfindings} indicates an NMT model by \citep{tang2020}, trained from mBART covering 50 languages. Shortnames: MBSE-G (greedy-selection),  MBSE-A (average-Smatch) `ens. dec.', ensemble decoding.}
\end{table*}


\begin{table*}[t]
    \centering
    \begin{tabular}{l|c|c|c|c|c|c|c|c|c} \hline
Languages & Smatch & Unlabeled & NoWSD & Concepts & NER & Neg. & Wiki & Reentrant & SRL \\\hline 
EN-mono  & \textbf{85.9} & \textbf{89.0} & \textbf{86.3} & \textbf{92} & \textbf{93} & \textbf{75} & \textbf{81} & \textbf{78} & \textbf{85} \\\hline
DE-cross & 73.7 & 77.9 & 73.8 & 75 & 89 & \textit{\textbf{48}} & 79 & 61 & 68 \\
ES-cross & 77.1 & 81.4 & 77.5 & 81 & 89  & \textit{\textbf{62}} & 79  & 67 & 74 \\
IT-cross & 76.1 & 80.4 & 76.3 & 79 & 90 & \textit{\textbf{56}} & 78 & 65 & 73 \\
ZH-cross & 63.0 & 67.9 & 63.1 & 65  & 85 & \textit{\textbf{35}} & 70 & 51 & 58 \\
\hline
DE-pipeline & 74.6 & 78.9 & 74.8 & 75 & 91 & \textit{\textbf{51}} & 80 & 62 & 68 \\
ES-pipeline & 77.1 & 81.1 & 77.3 & 80 & 91 & \textit{\textbf{61}} & 79 & 66 & 74 \\
IT-pipeline & 76.7 & 80.9 & 76.9 & 79 & 91 & \textit{\textbf{58}} & 80 & 65 & 73 \\
ZH-pipeline & 64.0 & 68.9 & 64.0 & 66 & 86 & \textit{\textbf{40}} & 74 & 51 & 59 \\
\hline
    \end{tabular}
    \caption{Fine-grained F1 scores on the AMR2.0 test set for EN (English), DE (German), ES (Spanish), IT (Italian) and ZH (Chinese). EN-mono denotes English mono-lingual parser, \{DE,ES,IT,ZH\}-cross, cross-lingual parsers and \{DE,ES,IT,ZH\}-pipeline, translate-and-parse pipeline.}
    \label{tab:finegrainedscore}
\end{table*}

\subsection{Cross-lingual AMR Parsing}
\label{crosslingual}

For cross-lingual AMR parsing, we consider the well known cross-lingual extension of AMR2.0 \cite{damonte-cohen-2018-cross}.
Our cross-lingual parsers are trained with Structured-mBART, always using separate vocabulary (sep-voc). 
Input sentences of the English training data are machine translated into the target languages with WLT\footnote{\url{https://www.ibm.com/cloud/watson-language-translator}} to generate cross-lingual parser training data. 

Table~\ref{tab:Results_Test2.0} shows the results on the human translated AMR2.0 test set, following standard practices. We provide results for recently published cross-lingual AMR parsers and different silver training versions of Structured-mBART. 
Structured-mBART with 4 model ensemble (Ensemble-4) distillation with just silver$^{1}$ improves the Smatch score by 2.1 to 2.6 over the Structured-mBART baselines, out-performing very strong previous SoTA from \citep{cai2021emnlpfindings} on Chinese and Spanish and tied on Italian. Increasing the input sentence diversity via AMR-to-text generation and ensemble decoding  further improve the system performances, attaining new cross-lingual SoTA on all four languages. Increasing the silver training data size to silver$^{1+2+3}$ and using 5 model ensemble (Ensemble-5) push the numbers higher by 0.2-0.5 Smatch points.

\cite{uhrig2021iwpt} report that translate-and-parse pipelines outperform the conventional cross-lingual parsers,  we thus include translate-and-parse from the combination of WLT and Structured-BART + MBSE distillation. This out-performs the cross-lingual parsers by 0.6-1.0 Smatch on all languages except for Spanish, when trained with the same MBSE avg.-Smatch silver$^{1+2+3}$ data.

Comparing the fine-grained F1 scores for cross-lingual parsers with those for English, as shown in Table~\ref{tab:finegrainedscore}, we observe that cross-lingual parsers are particularly worse than English for negation. For instance, German negations are often realized as a compound, as in \textit{\textbf{nicht}tarif{\"a}re} (non - tariff), which is aligned to the non-negated stem portion of the concept \textit{tariff}, losing its negation meaning. We observe similar issues in English with prefixed negations such as \textit{\textbf{un}happy, \textbf{in}adequate, \textbf{a}typical}.

\subsection{Domain Adaptation}
\label{domainadaptation}

We use the AMR2.0 version of BioAMR (medical domain) as this has clearly defined partitions\footnote{amr.isi.edu/download/2016-03-14/amr-release-training-bio.txt, amr-release-dev-bio.txt, amr-release-test-bio.txt}~and was used in  \citet{bevilacqua2021aaai}. We also use QALD-9-AMR, constructed from QALD-9 data\footnote{\url{https://github.com/ag-sc/QALD}} \cite{ricardo2018ceur-ws}, a corpus of natural language questions for executable semantic parsing \cite{kapanipathi-etal-2021-leveraging}.  Corpus statistics of the domain adaptation data is summarized in Table~\ref{tab:benchmarkdata}.

Table~\ref{tab:dadetailedsmatch} shows the experimental results.  Results for SPRING are taken from \citet{bevilacqua2021aaai}.
For each test set, we report the results under three different training conditions, all of which include either AMR2.0 or AMR3.0 treebank in the training data: (1) use only silver data with MBSE distillation, (2) use only domain gold sentences, (3) use both silver data and domain gold sentences. Since BioAMR is annotated in AMR2.0 style and QALD-9-AMR in AMR3.0 style, we use the corresponding Structured-BART models as indicated in the table.  

\begin{table}[!b]
\centering
\begin{tabular}{l|c|c} \hline
\textbf{Training Data} & \textbf{Smatch} & \textbf{NER} \\\hline
\hline
\multicolumn{3}{c}{\textbf{BioAMR Evaluations}} \\ 
\hline
SPRING$^{\mbox{DFS}}$ & 59.7 & \\
SPRING$^{\mbox{DFS}}$+ silver & 59.5 & \\
SPRING$^{\mbox{DFS}}$ (In domain) & 79.9 & \\
\hline
\textbf{Ours} & & \\
Struct-BART (AMR2.0)  & 60.4 & 27.0 \\
+MBSE-G silver$^{1}$ & 63.2 & 31.0 \\
+MBSE-G silver$^{b}$ & 66.9$\pm{0.2}$ & 30.0  \\
+MBSE-P silver$^{b}$ & 66.9$\pm{0.2}$ & 31.0 \\
\hline
+201 domain gold sent. & 72.3$\pm{0.2}$ & 68.0 \\
+403 domain gold sent. & 74.3$\pm{0.2}$ & 70.0 \\
+5K domain gold sent. & 79.8$\pm{0.2}$ & 80.0 \\
\hline
+MBSE-G silv.$^{b}$+201 gold & 75.8$\pm{0.3}$ & 70.0 \\
+MBSE-G silv.$^{b}$+5K gold & \textbf{81.3}$\pm{0.2}$ & \textbf{81.0} \\
\hline
\multicolumn{3}{c}{\textbf{QALD-9-AMR Evaluations}}  \\\hline
\textbf{Ours} & \\
Struct-BART (AMR3.0) & 87.2 & 84.0 \\
+MBSE-G silver$^{1}$ & 88.0 & \textbf{88.0} \\
+MBSE-G silver$^{q}$ & 89.5$\pm{0.1}$ & 85.0 \\
+MBSE-P silver$^{q}$ & 89.3$\pm{0.2}$ & 87.0  \\
\hline
+200 domain gold sent. & 88.5$\pm{0.5}$ & 84.0 \\
+408 domain gold sent. & 89.8$\pm{0.1}$ & 86.0 \\
\hline
+MBSE-G silv.$^{q}$+200 gold & 90.0$\pm{0.3}$ & 87.0 \\
+MBSE-G silv.$^{q}$+408 gold & \textbf{90.1}$\pm{0.1}$ & 87.0 \\
\hline
\end{tabular}
\caption{Smatch scores on BioAMR and QALD-9 test sets with varying sizes of human annotated (gold) domain sentences and silver data. MBSE-G (greedy-select) and MBSE-P (Graphene-Smatch respectively). MBSE distillations are all with Ensemble-4 (APT + Structured-BART + SPRING$_{1}$ + SPRING$_{2}$).}
\label{tab:dadetailedsmatch}
\end{table}

As for BioAMR data, MBSE distillation (with both graphene-Smatch and greedy-select) on silver$^{b}$ -- comprising PubMed (\textsc{LDC2008T20, LDC2008T21}), BioNLP-ST-2011 and CRAFT -- improves over the Structured-BART baseline by 6.5 Smatch point (60.4 vs. 66.9). However, adding just 201 domain gold sentences to AMR2.0 treebank results in 11.9 Smatch point improvement over the baseline (60.4 vs. 72.3). A close inspection shows that this is largely due to the NER score improvement, as shown in the column under \textbf{NER}, i.e. NER score 27.0 in Structured-BART (AMR2.0) vs. 68.0 after adding 201 domain gold sentences. The dramatic impact of NE coverage no longer holds when we double the domain gold sentences from 201 to 403. In fact, MBSE greedy-select silver$^{b}$ + 201 domain gold sentences (75.8) is more effective than doubling the domain gold sentences (74.3). Finally, by combining MBSE distillation on silver$^{b}$ with 5K domain gold sentences, the system achieves 81.3 Smatch, outperforming the previous SoTA by 1.4.

Regarding QALD-9-AMR data, MBSE distillation on silver$^{q}$, i.e. SQuAD-Q(uestion) sentences, is almost as effective as 408 domain gold sentences (89.8) for both graphene-Smatch (89.3) and greedy-select (89.5) algorithms.
Combining 408 domain gold sentences with MBSE greedy-select silver$^{q}$ adds less than 1 Smatch point to 90.1. 

Since MBSE distillation on silver$^{b}$ lags behind the performance of 201 human annotated AMR for BioAMR, mostly due to low NER scores, we further analyze the target vocabulary coverage of named entity (NE) types occurring in the test sets. The analysis is shown in Table~\ref{tab:netypeanalysis}. NE types are equally well covered in all models for Q9AMR (QALD-9-AMR). 0.7\% out-of-vocabulary (OOV) ratio is caused by a typo in human annotation of the test set, i.e. \textit{country} misspelled as \textit{countrty}. For BioAMR, however, NE type OOV ratio of MBSE silver model is 3.7\%, e.g. \textit{protein-segment}, \textit{macro-molecular-complex}, substantially higher than 0.6\% of the model trained with 201 domain gold sentences. When the NE type is OOV, there is no chance for the system to produce the missing NE type, let alone predicting it correctly, underscoring  the challenges posed by domain specific concepts unavailable elsewhere.

\begin{table}
    \centering
    \begin{tabular}{l|c|c} \hline
 Models            & BioAMR & Q9AMR \\\hline
 Structured-BART    &  8.7\%  &  0.7\% \\
 +MBSE silver     &  3.7\%  &  0.7\% \\
 +200 domain gold sents &  0.6\%  &  0.7\% \\\hline
    \end{tabular}
    \caption{Named entity (NE) type out-of-vocabulary ratio w.r.t the target vocabulary of various models. BioAMR and QALD-9 test sets include 1691 and 150 occurrences of named entities, respectively.}
    \label{tab:netypeanalysis}
\end{table}
\section{Related Work}
\label{relatedwork}

There have been numerous works applying ensemble/knowledge distillation \cite{hinton2015distilling} to machine translation \citep{kim2016emnlp, freitag2017arxiv, nguyen2020neurips, wang2020aaai, wang2021acl}, dependency parsing \citep{kuncoro2016emnlp} and question answering \citep{mun2018neurips, ze2020wsdm, you2021icassp, chen2021www}. 
Regarding ensembling AMR graphs, \citet{barzdins2016} propose choosing the AMR with highest average sentence Smatch to all other AMRs. \citet{graphensemble2021} proposed a more complex technique capable of building new AMRs by exploiting Smatch's hill climbing algorithm. Our work brings together ensemble distillation and Smatch-based ensembling and shows that it can provide substantial gains over the standard self-training. 

\citet{damonte-cohen-2018-cross} show that it may be possible to use the original AMR annotations devised for English as representations of equivalent sentences in other languages. 
\citet{damonte-cohen-2018-cross, youngsuk2021eacl} propose annotation projection of English AMR graphs to target languages to train cross-lingual parsers, using  word alignments. \citet{blloshmi-etal-2020-xl} show that one may not need alignment-based parsers for cross-lingual AMR, and model concept identification as a \textit{seq2seq} problem. \citet{sgl2021naacl} reframe semantic parsing as multilingual machine translation (MNMT) and propose a seq2seq architecture fine-tuned on pretrained-mBART with an MNMT objective. 
\citet{cai2021aclfinding} propose to use bilingual input to enable a model to predict more accurate AMR concepts. \citet{xlpt2021acl} propose a cross-lingual pretraining approach via multitask learning for AMR parsing.
\citet{cai2021emnlpfindings} propose to use noisy  knowledge distillation for multilingual AMR parsing. 
We introduce Structured-mBART and attain new SoTA in Chinese, German, Italian and Spanish cross-lingual parsing by applying MBSE distillation and AMR-to-text.

We subsume domain adaptation under data augmentation with MBSE distillation, where the only difference between the two lies in the properties of the unlabeled data. The unlabeled data is drawn from the target domain for the purpose of domain adaptation rather than those similar to the source training data for data augmentation in general.

\section{Conclusion}
\label{conclusion}

We proposed a technique called Maximum Bayes Smatch Ensemble (MBSE) distillation, which brings together  Smatch-based model ensembling  \citet{barzdins2016,graphensemble2021} and ensemble distillation \citet{hinton2015distilling} of heterogeneous parsers, to significantly improve AMR parsing. The technique generalizes well across various tasks and is highly effective, leading to a new single system SoTA in English and cross-lingual AMR parsing and achieving the performance comparable to human annotated training data in domain adaptation of QALD-9-AMR corpus. Remaining technical challenges include tokenization and alignment of an input token corresponding to more than one concept for AMR parsing and identification of unknown named entities and their types for domain adaptation.



\bibliographystyle{acl_natbib}
\bibliography{revisedpaper}

\newpage

\appendix

\section{Greedy-Select Ensemble Algorithm}
\label{sec:appendix_algorithm}

\begin{algorithm}[h]
\caption{Greedy-Select MBSE Algorithm and Corpus Selection}
\label{alg:algorithm1}
\textbf{Input}: AMR$_{1}$...AMR$_{n}$ parses from $n$ AMR parsing models, where $n \geq 3$ \\
\textbf{Optionally Require}: Smatch score threshold = $\theta$ \\
\textbf{Output}: One-best AMR parse

\begin{algorithmic}[1] 
\STATE Let bestAMR = $NULL$
\vspace{1mm}
\FOR{$\forall_{i, j}$ in $1 \leq i,j \leq n$ and $i \neq j$}
     \STATE Compute sentence Smatch score  $smatch$(AMR$_{i}$, AMR$_{j}$), total $n(n-1)/2$ scores.
     \STATE Pick the highest $smatch$(AMR$_{i}$, AMR$_{j}$).
       \FOR{Each AMR$_{a}$, where $a=i$ or $a=j$}
         \STATE Pick the highest $smatch$(AMR$_{a}$,  AMR$_{b}$)
          \IF{$a=i$}
           \STATE bestAMR = AMR$_{i}$
          \ELSE
             \STATE bestAMR = AMR$_{j}$
          \ENDIF
      \IF{$smatch$(AMR$_{a}$,AMR$_{b}$) $<$ $\theta$}
        \STATE bestAMR = $NULL$ 
        \STATE \COMMENT{no AMR to be used from this sentence}
      \ENDIF
    \ENDFOR
   \ENDFOR
\vspace{1mm}
\STATE \textbf{return} bestAMR
\end{algorithmic}
\end{algorithm}

We start with $n$ parses from $n$ heterogeneous parsing models, where the minimum number of parses is $3$. For each input sentence, we compute sentence-level Smatch scores between any two parses across all $n$ parses, for a total of $n(n-1)/2$ Smatch scores (lines 2-3). Subsequently, we pick the two parses AMR$_{i}$ and AMR$_{j}$ with the highest Smatch score, where AMR$_{i}$ denotes the AMR parse from the system ${i}$ (line 4)
For each of the two parses, AMR$_{i}$ and AMR$_{j}$, we choose the parse with the higher Smatch score against the rest of the parses as the best parse (lines 5-11). When the scores are tied, we select the first parse output (equivalent to a random choice of fixed seed).
 
We incorporate an optional parse selection criterion into Algorithm~\ref{alg:algorithm1},  indicated as \textbf{Optionally Require} and specified in lines 12-15. The bestAMR for input sentence is selected for data augmentation if the Smatch score $smatch$(AMR$_{a}$,  AMR$_{b}$) is greater than or equal to the pre-specified value $\theta$. 

\begin{table}[!t]
\centering
\begin{tabular}{l|r|r} \hline
Models   & AMR2.0 & AMR3.0 \\
\hline
Structured-BART & 34,156 & 33,200 \\
SPRING$_{1}$ & 25,129 & 29,407 \\
SPRING$_{2}$ & 17,866 & 17,830 \\
APT          &  10,235 & 6,949 \\
Total        & 87,386 & 87,386 \\
\hline
\end{tabular}
\caption{Distribution of  individual model parses from MBSE greedy-select distillation with silver$^{1}$ dataset in Table~\ref{tab:benchmarkdata}}
\label{tab:distribution}
\end{table}

\section{Model Structures and Parameter Size}
\label{sec:appendix_param}

Pre-trained BART and mBART share the same model configurations except for the vocabulary size. There are 12 encoder/decoder layers, 16 heads per layer, 1024 model dimension and 4096 feed forward network (FFN) size. BART includes \textasciitilde{50K} and mBART, \textasciitilde{250K} task vocabulary.

When using separate vocabulary (sep-voc), Structured-BART and Structured-mBART use the same vocabulary as BART and mBART, respectively, for the source. For the target,  they create embedding vectors for action symbols and the target vocabulary size vary according to the training data. When using joint vocabulary (joint-voc), Structured-BART shares the same vocabulary between the source and the target, a combination of BART vocabulary and the additional embedding vectors for some action symbols.

Vocabulary and parameter sizes for Structured-BART and Structured-mBART trained with MBSE  distillation are shown in Table~\ref{tab:structbartvoc} and Table~\ref{tab:structmbartvoc}, respectively.

\begin{table}
    \centering
    \small
    \begin{tabular}{l|c|c|c} \hline
Model & Param & AMR2.0 & AMR3.0 \\\hline
      & src voc size & 50,265 &  50,265   \\
sep-voc & tgt voc size & 42,344 & 42,784   \\
      & \# param  & 493,011,968  & 493,913,088  \\\hline
joint-voc & joint voc size & 57,912 & 58,673   \\
      & \# param  & 414,121,984 &  414,901,248 \\\hline
    \end{tabular}
    \caption{Vocabulary and parameter sizes of Structured-BART with MBSE  distillation on silver$^{1+2+3}$ dataset from Table~\ref{tab:benchmarkdata}}
    \label{tab:structbartvoc}
\end{table}

\begin{table}
    \centering
    \begin{tabular}{l|c|c} \hline
    Languages & voc size & \# param \\\hline
    DE (German) & 34,689 & 681,894,912 \\
    ES (Spanish) & 34,881 & 682,288,128 \\
    IT (Italian) & 33,681 & 679,830,528 \\
    ZH (Chinese) & 59,473 & 732,652,544 \\\hline
    \end{tabular}
    \caption{Target vocabulary (sep-voc) and parameter sizes of Structured-mBART with MBSE distillation on silver$^{1+2+3}$ dataset from Table~\ref{tab:benchmarkdata}. Source vocabulary size is 250,027 across all languages.}
    \label{tab:structmbartvoc}
\end{table}

\section{Implementation Details}
\label{sec:appendix_implementation}

\begin{table}[!t]
    \centering
    \begin{tabular}{l|c|c|c} \hline
    Lgs. & vocab & base model & ens. model  \\\hline
    EN & joint-voc & 60min & 60min \\
    DE & sep-voc & 23min & 42min \\
    ES & sep-voc & 24min & 44min \\
    IT & sep-voc & 22min & 40min \\
    ZH & sep-voc & 30min & 60min \\\hline
    \end{tabular}
    \caption{Inference time for AMR2.0 test set. Base models are trained on AMR2.0 treebank only and ens. models are trained on AMR2.0 treebank plus silver$^{1+2+3}$.}
    \label{tab:inferencetime}
\end{table}
Our models are implemented with 
\textsc{FAIRSEQ}
toolkit \citep{ott2019fairseq}, trained and tested on a single NVIDIA Tesla A100/V100 GPU with 40-80GB memory. We use fp16 mixed precision training and all models are trained on 1 GPU.

For all English AMR parsing models with silver data, we use the Adam optimizer with $\beta_1=0.9$ and $\beta_2=0.98$. Batch size is set to 1024 maximum number of tokens with gradient accumulation over 8 steps. Learning rate schedule is the same as \citet{vaswani2017} with 4000 warm-up steps and $1\mathrm{e}{-7}$ warm-up initial learning rate and  the maximum learning rate  $1\mathrm{e}{-4}$. Dropout rate is 0.2 and label smoothing rate is  0.01. These hyper parameters are fixed and not tuned for different models and datasets. All models are trained for 10 epochs and the best 5 checkpoints are selected based on the development set Smatch from greedy decoding. Model parameters are averaged over the top 3 and top 5 models. The model that produces the highest development set score, after beam search decoding with beam size = 1, 5 and 10, is selected as the final model.
Training with MBSE greedy-select silver$^{1+2+3}$ takes 48-72 hours, and all other models with less silver data take less time to train.

For cross-lingual AMR parsing, maximum learning rate is always set to $3\mathrm{e}{-5}$. Baseline models trained only on AMR2.0 corpus are trained up to 80 epochs whereas models with silver$^{1}$ (and AMR-to-text) is trained up to 30 epochs and models with silver$^{1+2+3}$, up to 15 epochs. Model parameters are updated after gradient is accumulated for 8192 tokens. Dropout rate, label smoothing rate and model selection criteria are the same as the English parsers. Training baseline models takes about 10 hours. Training with silver$^{1}$ takes about 24 hours. Training with silver$^{1+2+3}$ takes about 96-120 hours. In order to reduce the vocabulary size, which subsequently reduces the model parameter size and memory requirement, we prune out singleton target vocabulary for training with silver data.

Inference time for AMR2.0 benchmark test set is shown in Table~\ref{tab:inferencetime}, where beam size=10 and batch size=64 for all languages.  EN is decoded on NVIDIA Tesla A100 and all other languages, on NVIDIA Tesla V100.

\end{document}